\documentclass[conference,a4paper,flushend]{iaria} 
\usepackage{amsmath}
\usepackage{minted}
\usepackage{multirow}
\usepackage{multicol}
\usepackage[most]{tcolorbox}
\usepackage[table]{xcolor}
\usepackage{booktabs}
\usepackage{fancyhdr}
\title{A Regression Framework for Understanding Prompt Component Impact on LLM Performance}
\author{
  \IEEEauthorblockN{%
    Andrew Lauziere, Jonathan Daugherty and Taisa Kushner}
  \IEEEauthorblockA{%
    Galois, Inc. \\
    {\{\tt{lauziere}, \tt{jtd}, \tt{taisa}\}@galois.com}
} }
\begin{document}

\fancyhead{}
\fancyfoot{} 

\pagestyle{fancy}
\fancyhead[L]{}
\fancyhead[R]{\thepage}
\fancyfoot[L]{\centering \scriptsize{This material is based upon work supported by the Defense Advanced Research Projects Agency (DARPA) under Contract No. HR001124C0319. Any opinions, findings and conclusions or recommendations expressed in this material are those of the author(s) and do not necessarily reflect the views of the Defense Advanced Research Projects Agency (DARPA).}}
\fancypagestyle{firstpage}{%
  \lhead{}
  \rhead{}
  \lfoot{ \centering \scriptsize{This material is based upon work supported by the Defense Advanced Research Projects Agency (DARPA) under Contract No. HR001124C0319. Any opinions, findings and conclusions or recommendations expressed in this material are those of the author(s) and do not necessarily reflect the views of the Defense Advanced Research Projects Agency (DARPA). Distribution Statement “A” (Approved for Public Release, Distribution Unlimited).}}
}
\fancyhead[R]{\thepage}

\maketitle

\thispagestyle{firstpage}

\begin{abstract}

As large language models (LLMs) continue to improve and see further integration into software systems, so does the need to understand the conditions in which they will perform. We contribute a statistical framework for understanding the impact of specific prompt features on LLM performance. The approach extends previous explainable artificial intelligence (XAI) methods specifically to inspect LLMs by fitting regression models relating portions of the prompt to LLM evaluation. We apply our method to compare how two open-source models, Mistral-7B and GPT-OSS-20B, leverage the prompt to perform a simple arithmetic problem. Regression models of individual prompt portions explain 72\% and 77\% of variation in model performances, respectively. We find misinformation in the form of incorrect example query-answer pairs impedes both models from solving the arithmetic query, though positive examples do not always improve performance. In terms of text-instructions, we find significant variability in the impact of positive and negative instructions -- these prompts have contradictory effects on model performance. The framework serves as a tool for decision makers in critical scenarios to gain granular insight into how the prompt influences an LLM to solve a task.
\end{abstract}

\section{Introduction}

Large language models (LLMs) are a class of deep neural networks which proficiently model patterns found in text data \cite{zhao_survey_2025}. These models have achieved dominant performance across tasks such as text summarization \cite{basyal_text_2023} and machine translation \cite{zhu_multilingual_2024}; their performance continues to improve as training corpora and learning strategies such as reinforcement learning develop \cite{brown_language_2020, ouyang_training_2022}. The models serve as the intelligence behind tools which have experienced cultural impact unseen in machine learning technology, such as \textit{Chat-GPT} \cite{noauthor_chatgpt_nodate}, \textit{Grammarly} \cite{noauthor_free_nodate} and \textit{Microsoft Copilot} \cite{efrene_what_nodate}. 

Despite comprehensive adoption, LLMs are some of the most opaque of machine learning methods. This is true for both intrinsic model features such as complexity due to size and inter-connectivity, as well as extrinsic aspects including the use of large and closed-source training corpora and optimization routines. These factors lead to to a concerning gap between deployment of LLMs and their understanding. Larger and more powerful models are being increasingly integrated into software systems as they find new use-cases and improve upon existing ones. However, these models are simultaneously becoming harder to understand and thus trust as they are less interpretable and more likely to be closed-source. 

LLMs and tools which leverage them are also highly influenced by the user-specified prompt. While guardrails imposed during fine-tuning, e.g., \cite{ouyang_training_2022}, can reduce risk, the models can still output unwanted or harmful content. This direct-access paired with a lack of understanding of the mechanisms in which the prompt drives LLM output compounds risk. Researchers and decision-makers alike are interested in uncovering how LLMs function. At the moment there is little understanding as to when, why, or how an LLM will perform on a given task; researchers point to high performance on benchmark datasets as reason to trust (and thus use) their models \cite{openai_gpt-4_2023, team_gemini_2025}. In response, some researchers argue for slowing down AI research \cite{noauthor_pause_nodate}, while others pursue developing technologies to audit it.

In this work, we help fill the gap in understanding how LLMs function by developing an explainable artificial intelligence (XAI) framework for analyzing the impact of prompting on LLM performance. We coin this framework \textbf{IAMs}: \textbf{I}nterpretable \textbf{A}ttribution \textbf{M}odel\textbf{s}. \textbf{IAMs} adapts previous XAI methods \cite{ribeiro_why_2016, mohammadi_explaining_2024} to fit regression models comprising features representing portions of the LLM input prompt. The framework enables granular and statistically rigorous inspection of LLM behavior by estimating the impact of various prompt components, and combinations thereof, on model performance via regression models.

Section~\ref{sect:RW} first introduces necessary background concerning LLMs and prompting followed by an overview of XAI and a review of research related to this work. Then, Section~\ref{sect:M} details the \textbf{IAMs} framework. Section~\ref{sect:R} applies \textbf{IAMs} to analyze and compare how two open-source LLMs solve an arithmetic problem. Finally Section~\ref{sect:C} summarizes findings and outlines future work. 

\section{Related Work} \label{sect:RW}

LLMs are unique from other data-driven models in that they are not only trained offline (i.e., unsupervised pre-training and supervised fine-tuning) as in traditional statistical learning, but they also use ``in-context learning'' via the prompt to guide the model during inference \cite{brown_language_2020}. As a result, the end-user has direct impact on the quality of the output, so much so that a common step of model training is to censure its ability to output harmful content \cite{ouyang_training_2022}. Designing the prompt, or \textit{prompting}, has become an integral part in effectively using an LLM.

Explainable artificial intelligence (XAI) methods enable interpretation of low-bias high-variance machine learning models; here, we focus on applying \textit{local model-agnostic} methods \cite{molnar_interpretable_2022} to understand LLMs by mutating the input prompt. Local methods aim to explain complex model behavior at a specific input, whereas model-agnostic tools demonstrate interpretability independent of the complex model at hand. Ribeiro et al. developed LIME: Local Interpretable Model-agnostic Explanations \cite{ribeiro_why_2016}, a surrogate modeling approach which ``can be applied to any classifier.'' In LIME, the local model is trained using sampled points around a specific input in the complex model space; then, a binary vector is formed representing the inclusion or exclusion of certain features, such as words in a text model or pixels in an image model. An interpretable model is trained on the binary vectors and associated labels. The authors also propose a ``fidelity-interpretability'' tradeoff using an L1 regularization term (i.e., Lasso: Least Absolute Shrinkage and Selection Operator) on the local model fitting. A higher penalty will drive parameter estimates of less impactful features to zero. 

Shapley values are a canonical local model-agnostic approach; the concept arises from game theory in which features of a model are ``allocated payout'' according to each one's contribution in a prediction \cite{shapley_notes_1951}. Lundberg and Lee unified the surrogate modeling approach in LIME and Shapley values in their SHAP: Shapley Additive Explanations framework \cite{lundberg_unified_2017}. The authors show a unique set of equations for guaranteeing Shapley value properties in an additive attribution model. Lundberg and Lee provide approximation algorithms for SHAP whereas both Grah and Thouvvenot \cite{grah_projected_2020} and Štrumbelj and Kononenko \cite{strumbelj_explaining_2014} provide general approximation algorithms for Shapley values. Lundberg et al. then extended SHAP to tree-based methods such as decision trees, random forests, and gradient-boosted trees \cite{lundberg_consistent_2019, lundberg_explainable_2019}. The method was then improved and applied to explain AI models used in medicine via tree-based methods \cite{amoukou_accurate_2023}. 

Mohammadi applied Grah and Thouvenot's algorithm \cite{grah_projected_2020} to estimate Shapley values according to a vectorization of the prompt \cite{mohammadi_explaining_2024}. The method assessed the extent to which certain tokens impacted LLM performance. Liu et al. also applied Shapley values to quantify which prompts best contributed to performance by considering an ensemble of prompts and iteratively pruning lower performing entries \cite{liu_prompt_2024}. The approach used the \textit{Data Shapley} algorithm \cite{ghorbani_data_2019}.

\section{Methodology} \label{sect:M}

\begin{figure*}[htbp]
    \centering
    \includegraphics[width=\linewidth]{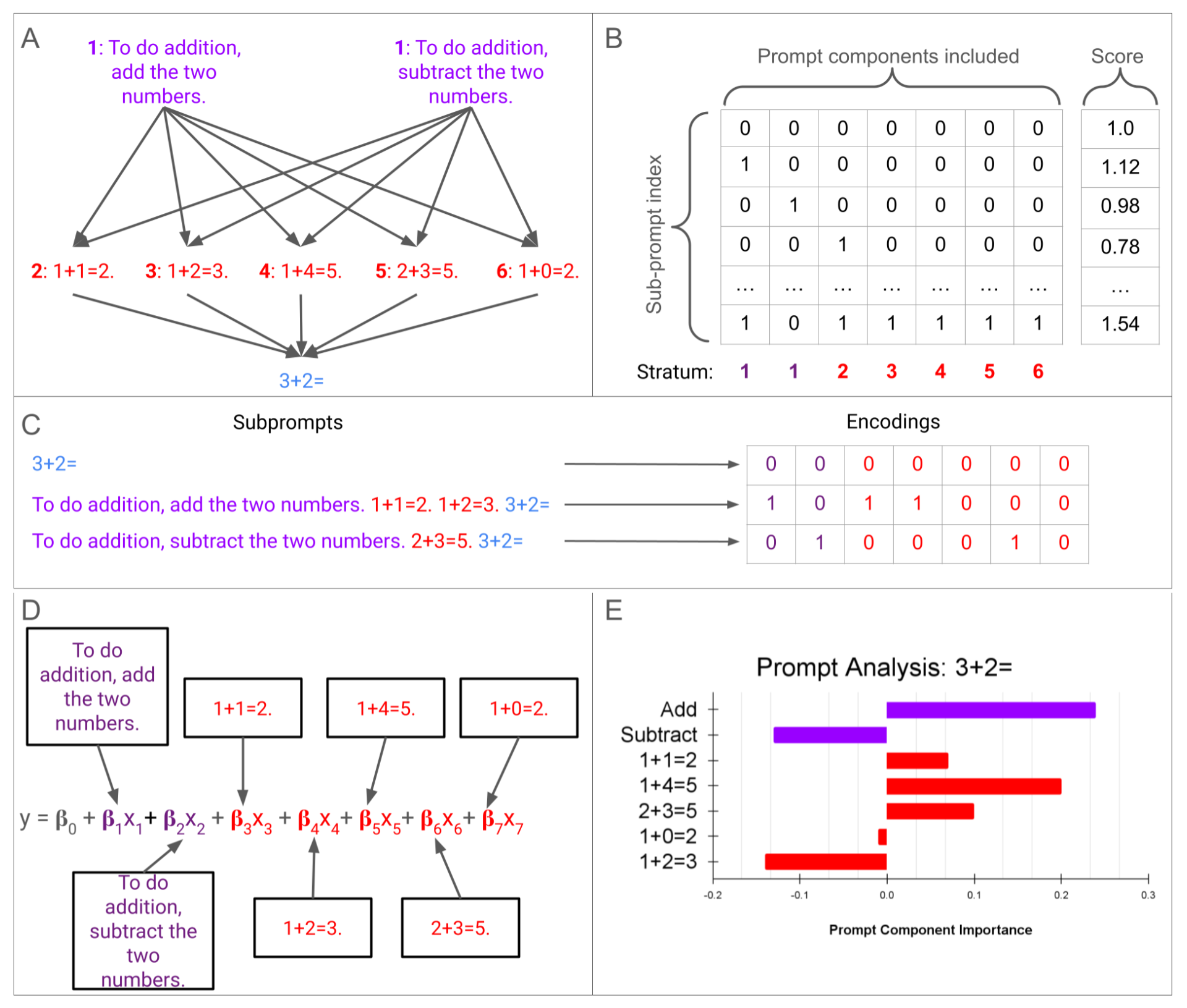}
    \caption{\textbf{Overview of IAMs.}  First in panel A, a prompt model describes texts of interest which could impact LLM performance on the query ``3+2='' (blue). The first stratum (purple) contains two choices of task-specific description texts while five strata (red) each have an example query-answer pair. An LLM processes and scores the full set of possible subprompts, each an input prompt containing a subset of prompt components (panel B). The inclusion (modeled with a 1) or exclusion (modeled with a 0) of components is expressed with binary variables (panel C). Pairs of encoding vectors and scores are used to fit a regression model, such as the one written in panel D. Panel E shows an example visualization of the learned model parameters conveying importance for LLM performance.}
    \label{fig:overview}
\end{figure*}

Figure~\ref{fig:overview} depicts an overview of \textbf{IAMs} applied to model how the prompt influences an LLM to solve an arithmetic problem. We outline individual steps here, and then describe details of each in the following subsections. 

First, panel A (top left) shows a prompt model specifying a query of interest (blue), ``3+2=,'' as well as other prompt components in six separate strata: one for task-specific text containing two choices (purple, numbered 1) and five example query-answer pairs each in an individual stratum (red, numbered 2-6). A sequence of prompt components, at most one from each stratum, forms a \textit{subprompt}. All subprompts arising from the possible inclusion or exclusion of each prompt component are processed by a chosen LLM, and the output is evaluated according to a user-specified score metric. Panel B (top right) shows the design matrix featuring seven binary variables, one for each prompt component, signaling the inclusion (or exclusion) of each component along with the corresponding output measurement score vector. Each prompt component is referenced in \textbf{bold}, corresponding to a column in the design matrix (directly under the matrix). Here, the score is a function of the probability of the leading output token corresponding to the correct answer, ``5.'' Scores could be binary, measuring whether or not the LLM answers a question correctly (e.g. returning ``5''), or continuous (e.g. the probability of the leading token corresponding to ``5''). Panel C (middle) shows how three example subprompts are encoded into binary vectors. The baseline subprompt, containing only the query of interest, is represented by an all-zero vector, as no other prompt components are used. Inclusion of at most one component from the first stratum is expressed in purple, while each example query-answer pair could be included, shown in red. The design matrix and score vector are used to fit a multiple regression model (D); visualizations allow insight into which prompt components drive LLM performance (E). 

\captionsetup{font={footnotesize,sc},justification=centering,labelsep=period}%
\captionsetup{font={footnotesize,rm},justification=centering,labelsep=period}%


\subsection{Prompt Stratification} \label{sect:PS}

Prompt stratification casts the prompt as a series of strata, in effect sets, anchored by a query of interest, such as a request or question. We decompose the prompt into \textit{m} sets, $\boldsymbol{p}_i$, $i = 1, 2, \dots, m$. Each prompt stratum has a selection of $n_i = |\boldsymbol{p}_i| \geq 1$ possible text choices (referred to as prompt components): $\{p_{i,1}, p_{i,2}, \dots, p_{i,n_i}\}$, where an element $p_{i,j}$ is text that could be included in the prompt at location \textit{i}. Each prompt stratum contains one (static) or multiple (variable) choices of text to include at that location in the prompt; the query stratum contains only the query itself. Permutations of all possible prompt components across strata yields the total subprompt set: \textit{S} = $\boldsymbol{p}_1 \times \boldsymbol{p}_2 \times ..., \boldsymbol{p}_m$ containing all $N = \prod_{i=1}^{m} |\boldsymbol{p}_i|$ unique prompts. A specified LLM processes all $N$ subprompts in \textit{S}. Texts contained \textit{within} a stratum are only seen one-at-a-time whereas texts \textit{between} strata will co-occur. Key here is the fact that co-occuring prompt components can be analyzed together in regression framework.
 
\subsection{Encoding Prompts as Binary Variables in a Regression Framework}

Our regression framework estimates both individual and joint effects of prompt components on LLM performance. Each unique subprompt contains the only component across static stratum (e.g. the query) and a subset of components across variable strata. Here, we detail steps to construct the binary design matrix from a prompt stratification such as that of Figure~\ref{fig:overview} panel A.

Denote the variable stratum $\varphi \subset \{1, 2, \dots, m\}$ with each $\boldsymbol{p}_l$, $l \in \varphi$, as as a categorical variable with $n_l+1$ levels. Implicitly, the empty string ``'' is included as a \textit{baseline} value in each variable stratum. A one-hot encoding maps the $n_l + 1$ levels to $(n_l + 1) - 1$ dummy variables, $j = 1, 2, \dots n_l$. We use the binary encodings

\begin{equation}
    x_{l, j, k} =
  \begin{cases}
                                   1 & \parbox{55mm}{if component $j$ in stratum $l$ is active in subprompt $k$} \\
                                   0 & \text{otherwise} 
  \end{cases}
\end{equation}

across variable stratum $l \in \varphi$ and levels $j=1, 2, \dots, n_l$ to estimate effects. For example, $p_1$ in Figure~\ref{fig:overview} panel A yields 2 dummy variables while each $p_2$, ..., $p_{6}$ generates one dummy variable each, totaling the seven columns in panel B. 

Then, interactions of dummy variables $x_{l,j,k}$ model the simultaneous occurrence of multiple prompt components across variable strata. A second-order interaction dummy variable, for $l' \neq l$ and $j' \in \{1, 2, \dots, n_{l'}\}$,

\begin{equation}
    x_{l, j, l', j', k} =
  \begin{cases}
                                   1 & \parbox{55mm}{if both component $j$ in stratum $l$ is active and component $j'$ in stratum $l'$ is active in subprompt $k$} \\
                                   0 & \text{otherwise} 
  \end{cases}
\end{equation}

captures subprompts with both $x_{l,j,k}=1$ and $x_{l', j',k} = 1$; prompt component interaction generalizes to \textit{L}-order ($2 \leq L \leq |\varphi|$) terms. 

Altogether, a selection of single dummy variables (i.e. first-order) and interaction terms (second-order, third-order, etc.) totaling \textit{M} variables produces a design matrix $\boldsymbol{X} \in\{0, 1\}^{(nN) \times (M+1)}$ akin to the matrix in Panel B of Figure~\ref{fig:overview}. The design matrix also includes an intercept term $\boldsymbol{1} = [1, 1, \dots, 1]' \in R^{nN}$ to estimate the score under the baseline value, i.e. when all other variables are 0. In context, this is the estimated score under the query alone with no other prompt components active. 

\subsection{Regression Modeling with Encoding Variables}

Each subprompt \textit{k}, represented by a row in the design matrix \textbf{X}, is processed by a chosen LLM. A user-specified scoring function evaluates model output and produces $y_k$. Continuous scores lead to a multiple regression approach whereas discrete (e.g. binary) output leads to a logistic regression fitting. \textbf{IAMs} supports continuous and binary measurements. The choice in a scoring function is limited by the LLM. Closed models only show the LLM output, as opposed to open models which reveal internal activations and token probability distributions. For example, when evaluating a closed model (e.g., GPT-5) the scoring function could evaluate the correctness of the output or inclusion of certain text (binary) or rate the output as in the application of LLMs-as-a-judge (continuous). On the other hand, open models such as GPT-OSS-20B can be evaluated using all closed model scoring functions, along with scores which measure model internals. Here in our demonstration (Section~\ref{sect:R}), we leverage model openness to use a higher information score than correctness: the probability value associated with the leading token being ``5.'' Broadly, continuous measurements such as ROGUE are commonly used to assess text summarization \cite{lin_rouge_2004}; Bilingual Evaluation Understudy (BLEU) is another continuous measurement to evaluate the accuracy of machine translation \cite{papineni_bleu_2002}.

Denote $\boldsymbol{y} = [y_1, y_2, \dots, y_{nN}]' \in R^{nN}$ as the real-valued measured output of all processed subprompts. A multiple regression of the form

\begin{equation} \label{eqn: unary_mlr}
    y_{k} = \beta_0 + \sum_{l \in \phi} \sum_{j=1}^{n_l} \beta_{l,j} x_{l,j,k} + e_{k}
\end{equation}

relates the binary $x_{l, j, k}$ variables to outputs $y_k$ via parameters $\beta_{l, j}$. The intercept parameter $\beta_0$ estimates the query score, serving as a baseline. Then, error terms $e_{k}$ are assumed to be independent and identically distributed: $N(0, \sigma^2)$. 

Interaction terms enable estimation of intra-stratum prompt component concurrence. Assume that two strata: \textit{l} and \textit{l'} are interacted, with $n_l$ and $n_{l'}$ sub-prompts in each, respectively. Then, up to $(n_l)*(n_{l'})$ interaction terms could be estimated: 

\begin{equation} \label{eqn: inter_mlr}
    y_{k} = \beta_0 + \sum_{l \in \phi} \sum_{j=1}^{n_l} \beta_{l,j} x_{l,j,k} + \sum_{j=1}^{n_l}\sum_{j'=1}^{n_{l'}} \beta_{l,j,l',j'} x_{l,j,k} x_{l',j',k} + e_{k}
\end{equation}

Similar to LIME, we include an L1 regularization term via $\lambda \in R^+$ in the fitting to pressure coefficients corresponding to less impactful prompt components (and interactions) to zero: 

\begin{equation} \label{eqn: l1}
    \hat{\boldsymbol{\beta}}_{\lambda} = \operatorname*{argmin}_{\boldsymbol{\beta}} \sum_{l \in \phi} \sum_{j=1}^{n_l} \sum_{k=1}^{nN} (y_{k} - \hat{y}_{k})^2 + \lambda |\boldsymbol{\beta}|
\end{equation}

Whereas the standard multiple regression optimization problem (Equation~\ref{eqn: unary_mlr}) has a closed-form solution, the L1-regularized least squares problem is solved via elastic-net, a coordinate descent scheme \cite{friedman_regularization_2010}. We use the \textit{statsmodels} implementation of elastic-net \cite{seabold_statsmodels_2010}.

The case in which $\boldsymbol{y} \in \{0, 1\}^{nN}$ comes about when the output measurement function tests a property (e.g. inclusion of information or factuality) of the LLM output. Prompt stratification and design matrix follow as in the continuous output measurement case above, but the L1-regularized logistic regression fitting is performed via SAGA \cite{defazio_saga_2014} implemented in scikit-learn \cite{pedregosa_scikit-learn_2011}. 

\subsection{Forward-selection Algorithm}

Model-selection algorithms proceed iteratively, adding terms from an empty model (i.e. forward-selection), removing terms from a full model (i.e backwards selection), or using a combination of both (bidirectional selection). Just as LASSO optimization (i.e. elastic-net \cite{friedman_regularization_2010}) reduces estimates without underlying context of their relationships, typical model selection algorithms do not assume any dependencies between variables. However, it is common practice when building interaction-term regression models to only include interaction terms if all corresponding subsets of terms are included in the model; e.g. both first-order variables in a second-order interaction term being present, and all three combinations of second-order interaction terms being present for a third-order interaction term. 

We adapt a forward-selection approach to only consider the incorporation of interaction terms in which lower level terms are already present. The algorithm starts with an intercept term and first iterates over first-order components; each is only incorporated based on the estimate's p-value relative to a Bonferroni corrected alpha level \cite{dunn_multiple_1961}. Then, interaction terms are considered up to a maximum level \textit{G} over a preselected subset of all interactions subject to increasingly higher thresholds. At each level $g = 1, 2, ..., G$, all interaction terms of that level among preselected strata are considered, subject to which (\textit{g-1})-level terms are already included. 

\subsection{Shapley Value Estimation}


We adjust the original Shapley value calculation \cite{molnar_interpretable_2022} to handle contributions of binary variables arising from a one-hot encoding of a categorical variable. This change is a result of the mutual exclusivity of binary variables of the same stratum; the original formula assumes any coalition of variables is feasible, yielding a variable-independent weighting of coalitions. Certain prompt components may appear in varying numbers of coalitions, depending on the prompt model at hand. 

Equation~\ref{eqn: shap_dummy} shows our updated formula taking into account the varying counts of coalitions when calculating the weighted average of marginal contributions. We follow the notation of \cite{molnar_interpretable_2022} though it conflicts with established \textbf{IAMs} notation. Here $k_i$ represents the total number of features that co-occur with prompt component \textit{i}, \textit{N} is the set of all prompt components, \textit{v} is the scoring function, and \textit{s} is a coalition among prompt components. In plain terms, the Shapley value of prompt component \textit{i} is the average marginal contribution of the component to all prompts it could be included within. The value is a weighted average of all such marginal contributions. 

\begin{equation} \label{eqn: shap_dummy}
    \varphi_i(v) = \frac{1}{k_i}\sum_{s \subset N \setminus \{i\}} (\frac{(k_i - |s|)!|s|!}{k_i!}
)^{-1} (v(s \cup \{i\}) - v(s)).
\end{equation}

\section{Comparing Mistral-7B and GPT-OSS-20B on Arithmetic} \label{sect:R}

Mistral-7B is a foundation model ``engineered for efficiency''; it outperformed open-source 13B and 34B models (Llama 2, and Llama 1, respectively) at time of publication across all evaluated benchmarks \cite{jiang_mistral_2023}. We chose this model to address two core concerns: first, it enables reproducibility by being open-source and small enough to be run on consumer hardware; second, it showed strong performance across multiple benchmark tasks such as GSM8K \cite{cobbe_training_2021} and MATH \cite{hendrycks_measuring_2021}. We compared Mistral-7B to GPT-OSS-20B, OpenAI's first open-source model since GPT-2 \cite{openai_gpt-oss-120b_2025}. Though much larger than Mistral-7B, GPT-OSS-20B's weights were quantized; according to \cite{openai_gpt-oss-120b_2025}, the model can be applied to systems with ``as little as 16 GB of memory.''

We applied our regression framework to inspect how both models processed the arithmetic query ``3+2='' by stratifying the prompt according to a choice of instruction texts and a set of examples. Our prompt stratification used $m=12$ strata with the last stratum containing the query alone: $p_{12}=\{\text{``3+2=''}\}$. We chose to conclude each prompt with the query as Mistral-7B is a foundation model; instruction texts and examples prior to the query were intended to guide the model to identify the token corresponding to ``5'' immediately after the query. 


The first stratum $\boldsymbol{p}_1$ contained $n_1 = 7$ choices of instruction text. The underscore token ``\_'' was added to measure the effect of ``token noise'' described in \cite{mohammadi_wait_2024} in which tokens containing low or unrelated information impact model performance. The next three instruction texts attempt to prime the models in performing arithmetic: ``Pretend you're a math expert.,`` ``To do addition, add the two numbers.,"'  ``To do subtraction, subtract the two numbers.'' The latter three instruction texts negate the three positive ones, respectively: ``Ignore what  I say next.,'' ``To do addition, subtract the two numbers.,'' ``To do subtraction, add the two numbers.'' This initial stratum, $\boldsymbol{p}_1$, enables measurement of token noise and how each model improves with positive instruction or is robust to negative instruction. Then, the next 10 stratum, $\boldsymbol{p}_2, ..., \boldsymbol{p}_{11}$ each contained an example-answer pair of similar arithmetic problems. The first five were all \textit{correctly} answered: ``0+1=1,"" ``1+1=2,'' ``1+2=3,'' ``2+3=5,'' ``1+4=5'' whereas the latter five were all \textit{incorrectly} answered: ``1+2=4,'' ``1+3=2,'' ``4+3=5,'' ``1+0=2,'' ``2+2=3.'' The inclusion of the incorrectly answered queries enabled insight into how each model is robust to misinformation.

In total, each model processed the same $8*2^{10}=8192$ unique prompts; there were eight choices in the first stratum (empty string and seven components) and each of the ten examples were either included or excluded (i.e. replaced with the empty string). We followed \cite{seals_evaluating_2024} in using Domain Conditional Pointwise Mutual Information (DCPMI) \cite{holtzman_surface_2021}. DCPMI weights token probabilities to avoid ``surface form competition,'' a phenomenon in which contextually similar tokens compete for probability mass \cite{holtzman_surface_2021}; this allowed us to process each prompt only once ($n=1$) using a one token generation step.

First, consider the correct token probability under the null prompt, i.e. \textit{Q} = ``3+2=.'' This baseline probability then serves as a reference point for the correct token probabilities arising from other subprompts. Let \textbf{y} be an output probability distribution over all tokens in the vocabulary and $y_c$ be the token corresponding to ``5,'' i.e. the ``correct'' token. The DCPMI of $y_c$ under the ``context'' of the query \textit{Q} when given the subprompt \textit{s} is defined

\begin{equation}
    DCPMI_Q(\mathbf{y}=y_c, s) = \frac{P(\mathbf{y}=y_c | s)}{P(\mathbf{y}=y_c | Q)}.
\end{equation}

Here, $P(\mathbf{y}=y_c | s)$ is the probability an LLM estimated ``5'' as the first token when given the subprompt \textit{s}, while $P(\mathbf{y}=y_c | Q)$ is estimated correct token probability when given the query alone. 
 
We followed the same fitting procedures for both Mistral-7B and GPT-OSS-20B (henceforth, Mistral and OSS, for brevity). An initial first-order multiple regression model was fit to establish a preliminary inspection for each model. Then, two interaction-term regression model sets were fit: the first set used an L1 regularization procedure across a grid of $\lambda$ values, while the second used our contributed forward-selection algorithm. The latter two sets of regression models contained interactions of examples, yielding insight into the marginal effects of in-context learning. 

\subsection{DCPMI Distributions}

\begin{figure}[htbp]
    \centering
    \includegraphics[width=.45\linewidth]{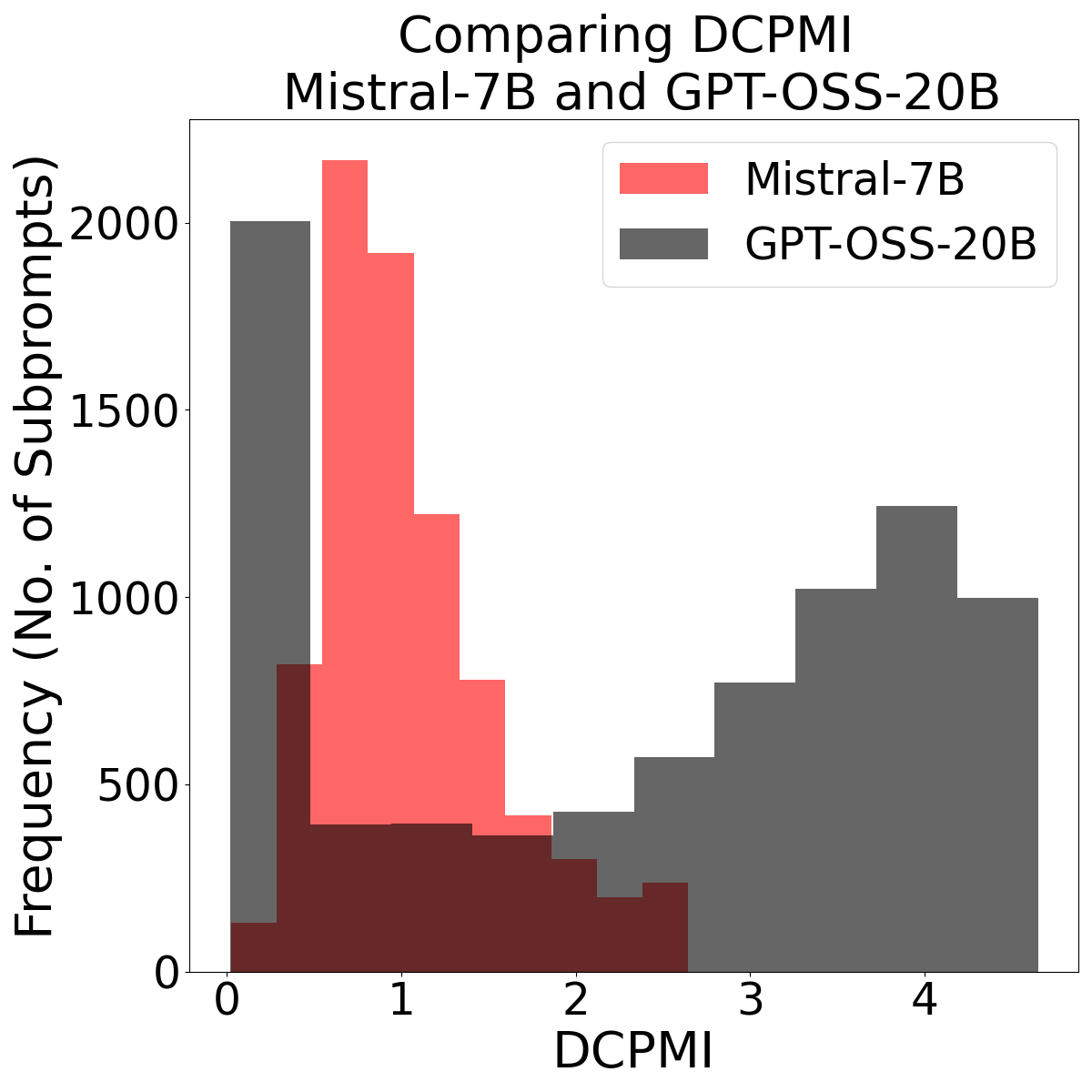}
    \includegraphics[width=.45\linewidth]{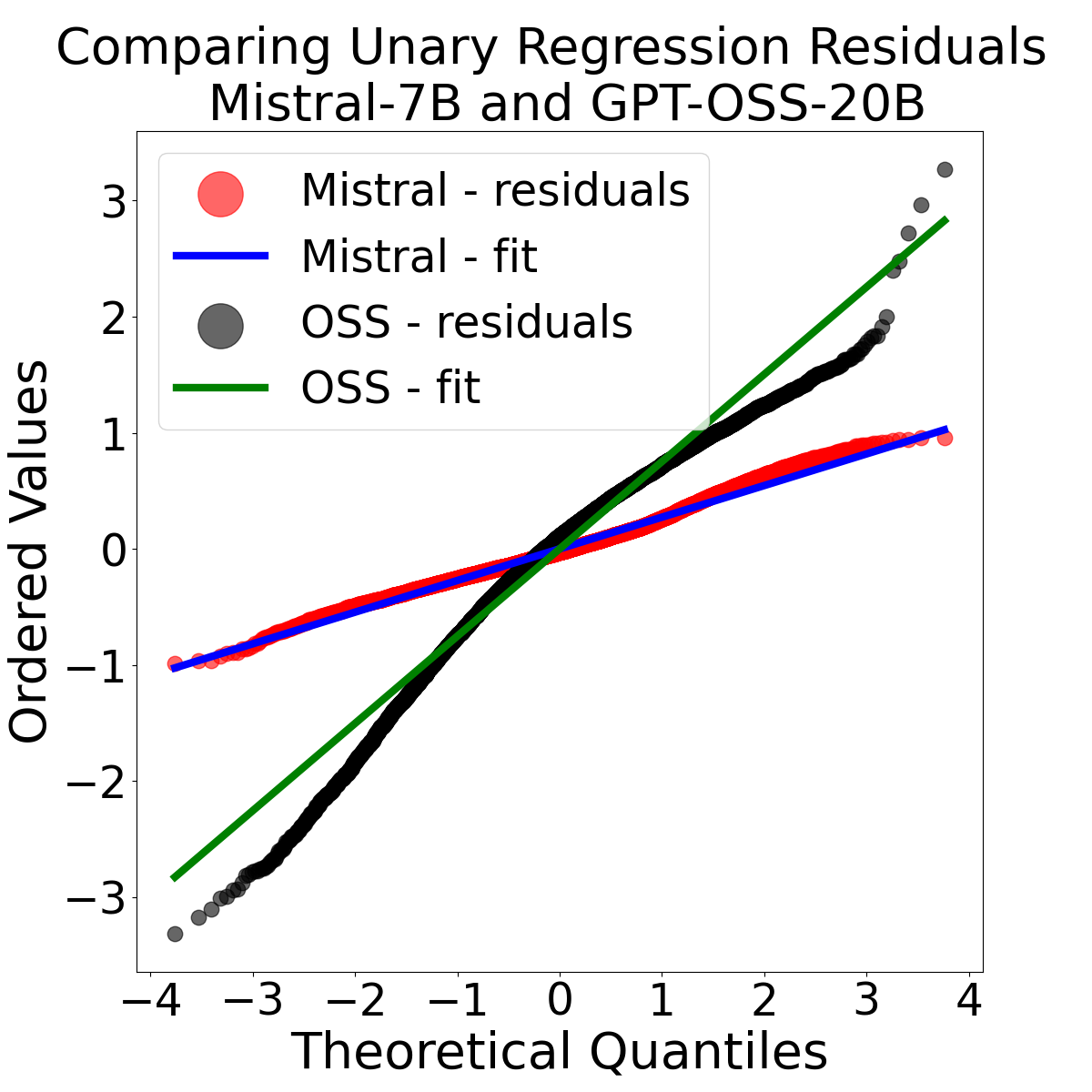}
    \caption{\textbf{GPT-OSS-20B was more sensitive to the prompt than Mistral-7B.} Left: Mistral-7B's DCPMI across all 8192 subprompts (red) showed a unimodal distribution whereas GPT-OSS-20B's DCPMI distribution was bimodal (black). The baseline DCPMI, 1.0 for both models, stands at the center of the Mistral-7B distribution. Right: Quantile-quantile (QQ) plots for Mistral (residuals in red, line of best fit in blue) and OSS (residuals in black, line of best fit in green) from first-order regression models. Overall, the unary model for OSS yielded more extreme residuals (y-axis values below -3 and above 3) than Mistral. The OSS residual distribution also showed deviation from normality as the tails of the distribution had less mass than expected (black dots under the green line).}
    \label{fig:dcpmi}
\end{figure}

Before investigating the regression results, we observed that the correct token probability under the query (i.e. baseline) was higher in Mistral than for OSS: 0.38 and 0.22, respectively. Mistral more effectively answered the query than the much larger and more recent OSS. However, we saw that the average DCPMI (i.e. the average of all correct token probabilities relative to the baseline probability) for OSS was over double that of Mistral, 2.36 to 1.05; this signaled that prompt components impacted OSS overall more positively than they did for Mistral. Figure~\ref{fig:dcpmi}'s left plot shows overlapping histograms of Mistral and OSSs' DCPMI distributions over all 8192 subprompts. The unimodality of Mistral's distribution and bimodality of OSS also suggested that OSS responded more to prompt components than Mistral. Positive components led to increased DCPMI while the negative ones were associated with reduced DCPMI in OSS, whereas the Mistral scores were anchored by the baseline (1.0) and fell slightly on either side of the center. 

\captionsetup{font={footnotesize,sc},justification=centering,labelsep=period}%
\captionsetup{font={footnotesize,rm},justification=centering,labelsep=period}%

\subsection{First-order Regression Models}

Initial first-order multiple regression models were fit using all 17 prompt components with an intercept term. The models showed how prompt components independently related to changes in DCPMI. First, we note the residuals in Figure~\ref{fig:dcpmi}'s right plot compares the distributions of residuals between models. OSS residuals (black dots) deviated slightly from normality (line of best fit in green), suggesting that the first-order model did not aptly explain DCPMI variation whereas the Mistral residuals (red dots) followed closer to the line of best fit (blue). First-order models had adjusted $R^2$ values of 0.72 and 0.76 for Mistral and OSS, respectively, which were surprisingly comparable given the the differences in observed DCPMI distributions. We then validated the regression framework by comparing Shapley values (Equation~\ref{eqn: shap_dummy}) for prompt components to first-order parameter estimates, finding Pearson correlation coefficients of 0.969 and 0.997, for Mistral and OSS, respectively, indicating a strong positive relationship and validating our modeling approach.

Table~\ref{tab:unary_reg_table} compares first-order regression coefficients between Mistral and OSS. Rows are color-coded according to sentiment: neutral in gray, positive or true in green, and negative or false in red. Asterisks correspond to p-values: *** for p < 0.0001, ** for p < 0.005, and * for p < 0.05. 

The ``Intercept'' coefficients (1.604 for Mistral and 3.324 OSS) estimated an average when all other variables are zero, which in context describes the baseline query DCPMI. Including other prompt components via binary variables then \textit{shifted} the estimated average. For example, when including the final prompt component ``2+2=3,'' (last row of Table~\ref{tab:unary_reg_table}) DCPMI lowered by 0.586 for Mistral and 0.144 for OSS, on average, all else constant. To contextualize the coefficients, view each as a change in percentage from the baseline query DCPMI: 36.5\% ($\frac{1.604 - 0.586}{1.604} = 0.635$) for Mistral, and 4.3\% ($\frac{3.324 - 0.144}{3.324} = 0.957$) for OSS. 

Contrary to expectation, the neutral token ``\_'' estimate had a large p-value for Mistral, 0.216, but a near-zero (statistically indiscernible from zero) p-value, less than 0.0001, for OSS. In context, Mistral was more robust to this the impact of this ``non-sense'' token than OSS, as the token estimate had a statistically significant negative (though small, -0.231) effect on DCPMI. This is notable as one might expect the larger, more recent model (OSS) to exhibit greater robustness than Mistral. 

Instruction texts showed incongruent effects between models, while example query-answer pairs aligned more with a 71\% Pearson correlation coefficient. Negative examples, in particular, hampered Mistral; e.g. ``1+3=2'' with estimate -0.450, ``1+2=4'' with estimate -0.270. On the other hand, instruction texts ``Pretend you're a math expert'' and ``Ignore what I say next'' negatively impacted OSS, with estimates -3.057 and -2.929. Prompt components overall hindered both models more than guiding them, as evidenced by summing all parameter estimates for each model, although true example query-answer pairs showed statistically significant gains to DCPMI each, on average. 

\captionsetup{font={footnotesize,sc},justification=centering,labelsep=period}%
\begin{table}[htbp]
\centering
\begin{center}
\begin{tabular}{c|c||c|}
 & Mistral \cite{jiang_mistral_2023} & OSS \cite{openai_gpt-oss-120b_2025} \\
 \textbf{Component} & \textbf{Coefficient} & \textbf{Coefficient} \\
\midrule
\textbf{Intercept}                                 &       1.604\textsuperscript{***}         &                3.324\textsuperscript{***}  \\
\hline
\rowcolor{lightgray}
\textbf{\_}                                           &      -0.020  &         -0.231\textsuperscript{***} \\
\hline
\rowcolor{lime}
\textbf{Pretend you're a ...}                &       0.045\textsuperscript{***}  &               -3.057\textsuperscript{***}   \\
\rowcolor{lime}
\textbf{To do addition, ...}         &       0.176\textsuperscript{***} &                0.130\textsuperscript{***}  \\
\rowcolor{lime}
\textbf{To do subtraction, ...} &      -0.037\textsuperscript{**}  &               -0.774\textsuperscript{***}  \\
\rowcolor{pink}
\hline
\textbf{Ignore what I ...}                      &      0.069\textsuperscript{***} &               -2.929\textsuperscript{***}          \\
\rowcolor{pink}
\textbf{To do addition, ...}    &       0.134\textsuperscript{***}  &                0.069\textsuperscript{*}    \\
\rowcolor{pink}
\textbf{To do subtraction, ...}      &       0.053\textsuperscript{***} &               -0.144\textsuperscript{***}       \\
\rowcolor{lime}
\hline
\textbf{0+1=1}      &     -0.033\textsuperscript{***}   &                0.538\textsuperscript{***}            \\
\rowcolor{lime}
\textbf{1+1=2}                                        &       0.033\textsuperscript{***}     &                0.210\textsuperscript{***}             \\
\rowcolor{lime}
\textbf{1+2=3}                                        &       0.062\textsuperscript{***}  &                0.153\textsuperscript{***}            \\
\rowcolor{lime}
\textbf{2+3=5}                                        &       0.228\textsuperscript{***}     &                0.127\textsuperscript{***}           \\
\rowcolor{lime}
\textbf{1+4=5}                                        &       0.111\textsuperscript{***}   &                0.343\textsuperscript{***}                \\
\rowcolor{pink}
\hline
\textbf{1+2=4}                                        &      -0.270\textsuperscript{***}   &               -0.277\textsuperscript{***}          \\
\rowcolor{pink}
\textbf{1+3=2}                                        &      -0.450\textsuperscript{***}   &               -0.745\textsuperscript{***}           \\
\rowcolor{pink}
\textbf{4+3=5}                                        &      -0.093\textsuperscript{***}   &               -0.085\textsuperscript{***}              \\
\rowcolor{pink}
\textbf{1+0=2}                                        &      -0.214\textsuperscript{***}   &               -0.305\textsuperscript{***}         \\
\rowcolor{pink}
\textbf{2+2=3}                                        &      -0.586\textsuperscript{***}   &               -0.144\textsuperscript{***}          \\
\end{tabular}
\end{center}
\caption{\textbf{Mistral-7B and GPT-OSS-20B similarly used example query-answer pairs to solve the query, but instruction texts derailed OSS.} Prompt components (\textbf{Component}) and corresponding first-order parameter estimates (\textbf{Coefficient}) for both Mistral and OSS explain how each model solved the query.  Positive or true components are marked in light-green background, negative or false components are marked with red background, and the neutral underscore is marked in gray. ***: p < 0.0001, **: p < 0.005, *: p < 0.05.}
\label{tab:unary_reg_table}
\end{table}
\captionsetup{font={footnotesize,rm},justification=centering,labelsep=period}%

\subsection{Regularized Regression Models}

We next fit a series of regularized regression models with interactions between example query-answer pairs up to the fourth degree, totaling 393 parameters: intercept, 17 first-order terms, ${10\choose 2}=45$ second-order interactions, ${10\choose 3}=120$ third-order interactions, and ${10\choose 4}=210$ fourth-order interactions. Both this set of models, and the upcoming forward-selection algorithm models, investigated the extent to which multiple examples, i.e. few-shot learning,  marginally impacted performance. We performed a grid search over values of $\lambda \in (0, 0.003]$ by increments of 0.00001 to identify an optimal value $\lambda^*$ and corresponding estimates $\hat{\boldsymbol{\beta}}_{\lambda^*}$ for both Mistral and OSS. The range was selected by iteratively halving the range from 1.0 to find the $\lambda$ values which caused variation in both the magnitude of parameter estimates and MSE.

\captionsetup{font={footnotesize,sc},justification=centering,labelsep=period}%
\begin{figure}[htbp]
\includegraphics[width=.9\linewidth]{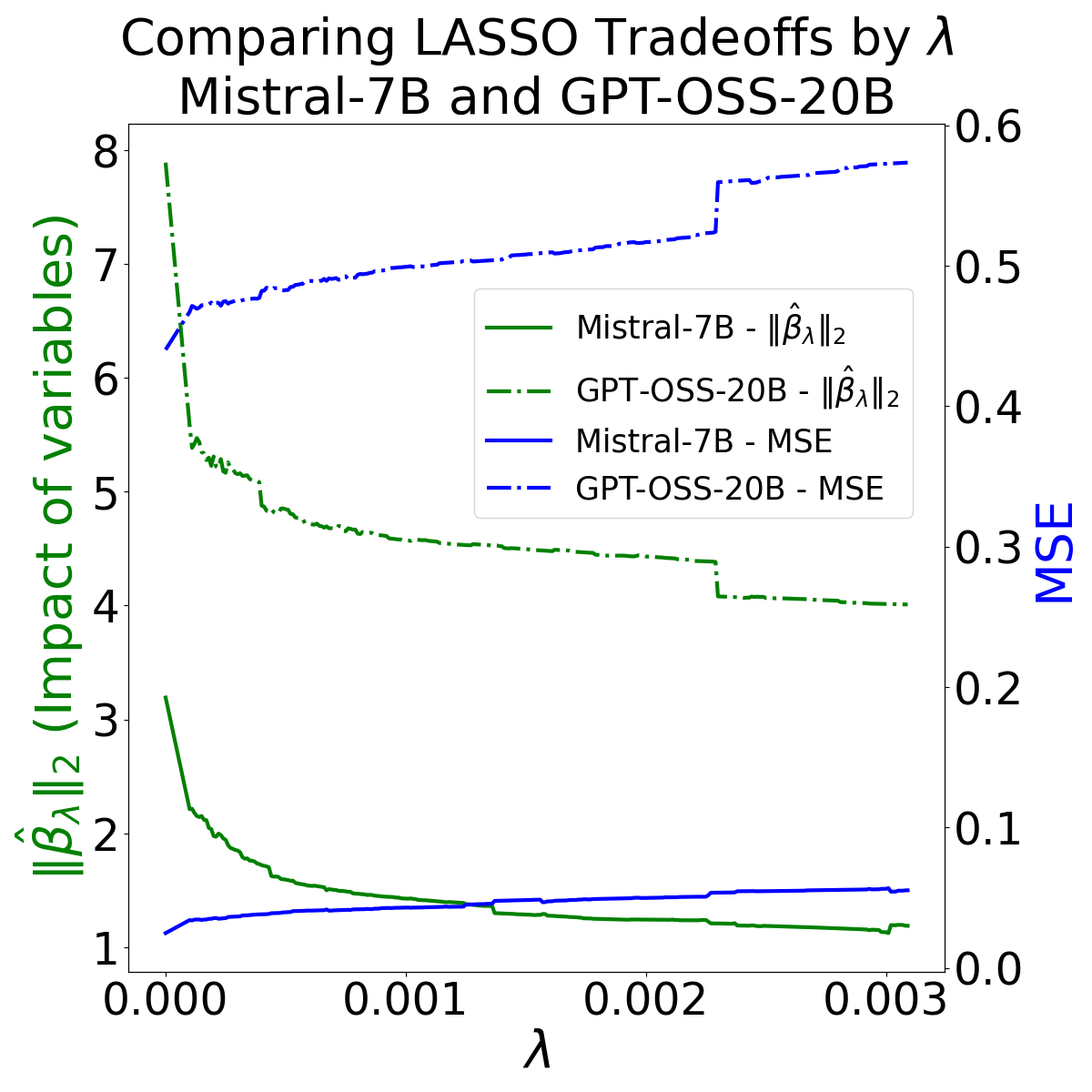} 
\caption{\textbf{LASSO trades off fidelity for interpretability.} LASSO penalized model size (green curves, measured $\|\hat{\boldsymbol{\beta}}_\lambda\|_2$) with increasing $\lambda$. As models shrank, performance lowered and MSE (blue curves) rose, yielding a trade-off. The grid search procedure identified an optimal $\lambda^* \approx 0.0004$ for both models. At these ``elbows'' of the green curves, model reduction began to slow.}
\label{fig:lasso-left}
\end{figure}
\captionsetup{font={footnotesize,rm},justification=centering,labelsep=period}%

Figure~\ref{fig:lasso-left} depicts how regularization impacted the interaction-term regression models. The plot shows model capacity, measured as the L2 norm of all parameter estimates ($\|\hat{\boldsymbol{\beta}}_\lambda\|_2$) against model performance, measured by mean-squared error (MSE). Green curves highlight model size (solid for Mistral, dashed for OSS) while blue curves depict MSE. The observed negative relationship follows from model regularization theory: constrained models are less able to capture variation and thus produce more error. Identifying the optimal $\lambda^*$ is selected at the ``elbow'' of the green curves, or where model size showed resistance to regularization, about 0.0004 for both models. Compared to the unregularized fitting (i.e. $\lambda$=0),  MSEs increased marginally while model sizes reduced by $\approx 40\%$ each (approximately 3.2 to 1.8 for Mistral and 7.9 to 4.8 for OSS). The optimal parameter estimation sets, $\hat{\boldsymbol{\beta}}_{\lambda^*}$, contained: 15 first-order, 37 second-order, 51 third-order, and 23 fourth-order nonzero terms in the Mistral model, and 16 first-order, 37 second-order, 70 third-order, 77 fourth-order nonzero terms in the OSS model. Selected regression models reported adjusted R-squared values 0.846 and 0.802, for Mistral and OSS, respectively. 

\captionsetup{font={footnotesize,sc},justification=centering,labelsep=period}%
\begin{figure}[htbp]
\includegraphics[width=.9\linewidth]{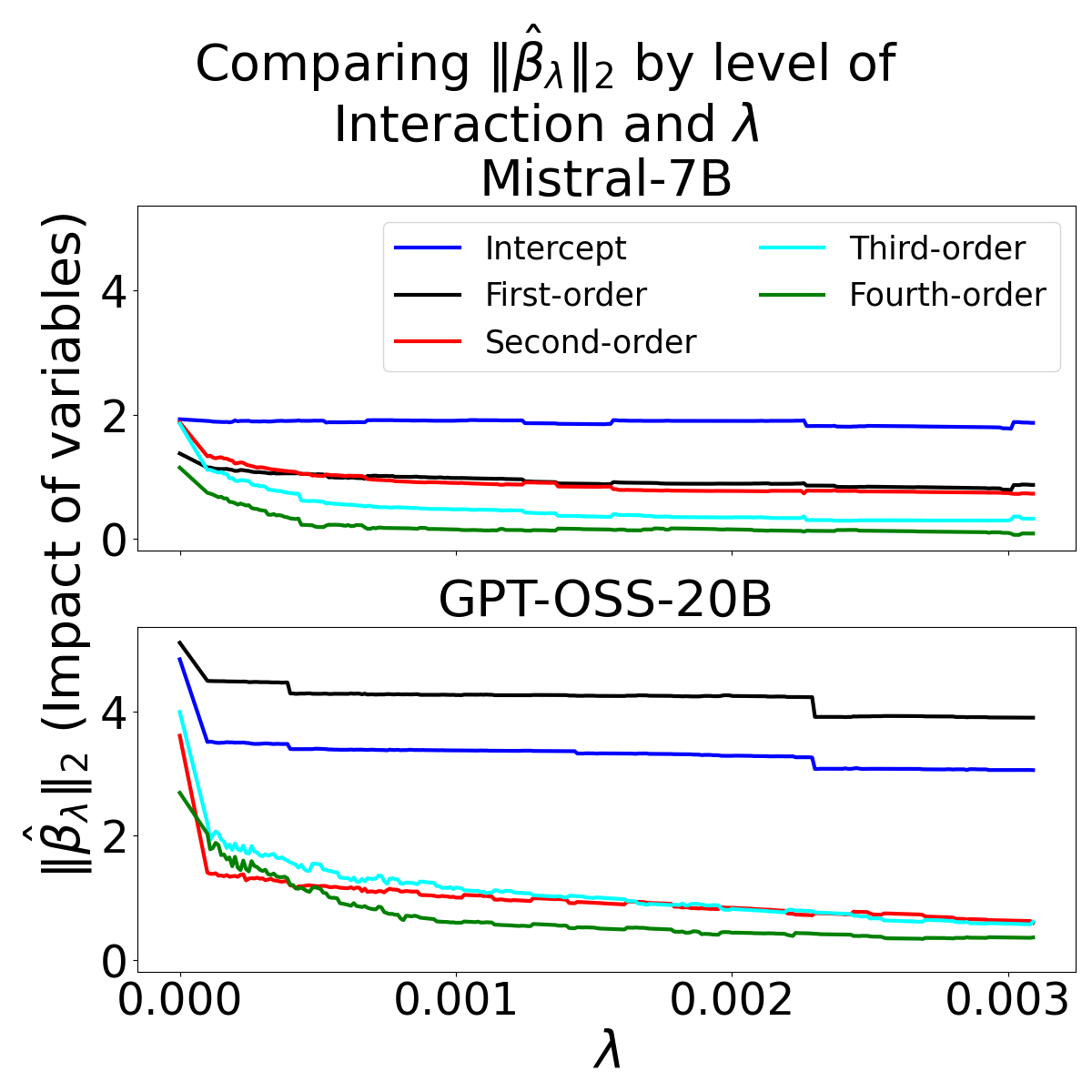} 
\caption{\textbf{LASSO penalized interaction terms more than first-order terms in both Mistral and OSS.} Model sizes, $\|\hat{\boldsymbol{\beta}}_\lambda\|_2$, were stratified by level of interaction. Blue curves show Intercept (baseline) estimates for each $\lambda$; OSS first-order estimates remained larger than the baseline whereas all prompt component parameter estimates fell below the baseline value for Mistral, suggesting that OSS used prompt components more than Mistral, relative to baseline query DCPMIs.}
\label{fig:lasso-right}
\end{figure}
\captionsetup{font={footnotesize,rm},justification=centering,labelsep=period}%

Then, Figure~\ref{fig:lasso-right} again shows Mistral (top) and OSS (bottom) parameter estimate norms, $\|\hat{\boldsymbol{\beta}}_\lambda\|_2$, but now stratified by order of interaction. Blue lines on each plot show the intercept estimate as a reference point for each level $\lambda$. A similar pattern is visible between models: lower-order terms showed stronger effects than increasingly higher order terms. However, first-order terms impacted OSS more than Mistral, relative to each set of baseline (Intercept) values.


\subsection{Forward-selection Models}

Finally, we applied our forward-selection procedure and identified a subset of the full 393-term regression model (up to fourth-order interactions among example query-answer pairs) for both Mistral and OSS starting with $\alpha$ = 0.05 and applying Bonferroni's correction at each level of interaction $g=1, 2, 3, 4$.  The resulting regressions achieved adjusted R-squared values of 0.861 and 0.779 using varying prompt components and prompt component interactions. Again, the individual incorrect query-example pairs showed strong negative effects on both models. Level-two interactions heavily favored false information in both models; query-answer pairs in which at least one of the two examples was false were overrepresented in the forward-selection models (18 of 19 in Mistral, 20 of 21 in OSS). Further, level-two interactions in which both examples were false tended to yield positive estimates, 11 of 14 in Mistral and 7 of 9 in OSS, signaling a softening effect of multiple pieces of misinformation. For example, in Mistral, the pairs: ``1+2=4'' and ``1+0=2,'' ``1+3=2'' and ``1+0=2,'' and ``1+0=2'' and ``2+2=3'' resulted in positive estimates: 0.261, 0.348, and 0.574. Then in OSS, the pairs: ``1+3=2'' and ``4+3=5,'' ``1+3=2'' and ``2+2=3,'' and ``4+3=5'' and ``2+2=3'' had estimates: 0.543, 0.330, and 0.346. Altogether, some false information disrupts both models, though decreasingly. 

\section{Conclusion \& Future Work} \label{sect:C}

\textbf{IAMs} provides a flexible approach to investigating prompt variation on LLM performance. Casting the prompt as a set of disjoint texts paved the way to modeling both individual, and coalitions of, prompt components via established frameworks such as LIME \cite{ribeiro_why_2016}. Individual and interaction regressions, regularized regression (LASSO), and our forward-selection algorithm jointly assist in identifying which, and to what extent, prompt components drive LLM performance. The approach was validated using an adjusted Shapley value calculation (Equation~\ref{eqn: shap_dummy}).

We applied the framework to inspect how the open-source models Mistral-7B and GPT-OSS-20B solved a single-digit arithmetic task, uncovering a vulnerability to misinformation (e.g. incorrect query-answer coefficients listed in the last five rows of Table~\ref{tab:unary_reg_table}), along with strong inconsistency in how text-based prompts (positive and negative) impacted performance. The toy-problem was simple enough to include two relied-upon strategies of prompt design: instructing and few-shot learning. We conclude that misinformation hampered both models more than correct information guided either, and that text-based prompts have unreliable effects on model performance, often having the opposite effect one would expect. Further experimenting in tandem with previous work on few-shot learning \cite{zhao_calibrate_2021, liu_how_2017, su_selective_2022} could uncover model-specific or task-specific intricacies of prompt design. 

Though not demonstrated, \textbf{IAMs} is applicable to closed-source models. Binary measurements of model output leads to a logistic regression approach; evaluations at multiple seeds gives a sense of correctness probability per subprompt. Subsequent analyses on prompt components, coalitions, and visualizations follow from the arithmetic demonstration here. 

We anticipate \textbf{IAMs} being especially helpful when LLMs are deployed in critical or high-risk scenarios; the framework will also enable granular insight into how in-context learning (i.e. prompting) impacts LLMs across scenarios. Furthermore, we see direct applications to evaluating and comparing AI agents. In particular, coding agents typically use a file which defines roles, personality, and other facets of the agent. Directly measuring how variations in this agent blueprint drive performance, or how agents compare on code generation more broadly, stands to improve such tools. Beyond coding agents, chatting agents which receive end-user evaluation could also be compared via \textbf{IAMs}.

In all, we believe the \textbf{IAMs} approach provides a rigorous framework for evaluating the impact of prompts on model performance, filling a gap in current literature in explainable artificial intelligence.

\printbibliography
\end{document}